\documentclass[a4paper, 10pt, conference]{ieeeconf}      % Use this line for a4
\usepackage{FG2026}

\FGfinalcopy % *** Uncomment this line for the final submission

\IEEEoverridecommandlockouts                              % This command is only
\usepackage{graphicx}
\usepackage{url}
\usepackage{amsmath,amssymb,amsfonts}
\usepackage{multirow}
\usepackage{booktabs}
\usepackage{caption}

\makeatletter
\let\NAT@parse\undefined
\makeatother

\definecolor{cvprblue}{rgb}{0.21,0.49,0.74}
\usepackage[colorlinks=true,linkcolor=cvprblue,citecolor=cvprblue,urlcolor=cvprblue]{hyperref}

\usepackage[sort,nocompress]{cite}
\def\FGPaperID{41} % *** Enter the FG2026 Paper ID here

\title{\LARGE \bf
TransFIRA: Transfer Learning for \\Face Image Recognizability Assessment
}

\author{\parbox{16cm}{\centering
    \vspace{-1.6mm}
    {\large 
    Allen Tu$^{1,2}$ \hspace{2em} 
    Kartik Narayan$^3$ \hspace{2em} 
    Joshua Gleason$^1$ \hspace{2em} 
    Jennifer Xu$^1$ \\ 
    Matthew Meyn$^1$ \hspace{2em} 
    Tom Goldstein$^2$ \hspace{2em} 
    Vishal M. Patel$^3$}\\
    {\normalsize
    $^1$ Systems and Technology Research \hspace{1em}
    $^2$ University of Maryland, College Park \hspace{1em}
    $^3$ Johns Hopkins University}
    \\ \url{https://transfira.github.io}
    }
    \vspace{-1.6mm}
}

\usepackage{fancyhdr}
\begin{document}

\ifFGfinal
\thispagestyle{plain}
\pagestyle{plain}
\maketitle
\else
\author{Anonymous FG2026 submission\\ Paper ID \FGPaperID \\}
\pagestyle{plain}
\maketitle
\fi

\thispagestyle{plain}

\begin{abstract}
Face recognition in unconstrained environments such as surveillance, video, and web imagery must contend with extreme variation in pose, blur, illumination, and occlusion, where conventional visual quality metrics fail to predict whether inputs are truly recognizable to the deployed encoder. Existing FIQA methods typically rely on visual heuristics, curated annotations, or computationally intensive generative pipelines, leaving their predictions detached from the encoder’s decision geometry. We introduce TransFIRA (Transfer Learning for Face Image Recognizability Assessment), a lightweight and annotation-free framework that grounds recognizability directly in embedding space. TransFIRA delivers three advances: (i) a definition of recognizability via class-center similarity (CCS) and class-center angular separation (CCAS), yielding the first natural, decision-boundary–aligned criterion for filtering and weighting; (ii) a recognizability-informed aggregation strategy that achieves state-of-the-art verification accuracy on BRIAR and IJB-C while nearly doubling correlation with true recognizability, all without external labels, heuristics, or backbone-specific training; and (iii) new extensions beyond faces, including encoder-grounded explainability that reveals how degradations and subject-specific factors affect recognizability, and the first method for body recognizability assessment. Experiments confirm state-of-the-art results on faces, strong performance on body recognition, and robustness under cross-dataset shifts and out-of-distribution evaluation. Together, these contributions establish TransFIRA as a unified, geometry-driven framework for recognizability assessment  that is encoder-specific, accurate, interpretable, and extensible across modalities, significantly advancing FIQA in accuracy, explainability, and scope.
% \\Project page and code: \url{https://transfira.github.io}
\end{abstract}
\section{INTRODUCTION}

Face recognition in unconstrained domains such as surveillance, long-form video, and web imagery must contend with severe variation in pose, blur, illumination, and occlusion. Under such conditions, recognition performance cannot be inferred reliably from superficial measures of image quality. What ultimately matters is whether samples are \emph{recognizable} to the deployed encoder, since even a few unrecognizable frames can corrupt template construction and compromise downstream matching. Bridging this gap between appearance-based quality and encoder-specific recognizability defines \emph{face image quality assessment (FIQA)} as a central challenge for robust recognition.

Despite progress in face recognition~\cite{wang2018cosface, deng2019arcface, narayan2025petalface, nair2025improved}, FIQA methods remain limited in generalization and interpretability. Hand-crafted cues such as sharpness, pose, or illumination~\cite{raghavendra2014automatic, lijun2019multi, gao2007standardization, terhorst2020ser, IconicityFace} often fail to reflect recognizability in the encoder’s embedding space. Regression- and uncertainty-based pipelines~\cite{hernandez2019faceqnet, SDD-FIQA2021, DBLP:conf/cvpr/TerhorstKDKK20} rely on curated labels, auxiliary supervision, or costly inference, making them indirect and resource-heavy. Model-based methods~\cite{meng2021magface, boutros2023cr, DBLP:conf/cvpr/KolfDB22} couple FIQA to specific backbones, restricting adaptability, while generative and multimodal pipelines~\cite{babnikIJCB2023, babnikTBIOM2024, Ou_2024_CVPR} improve performance but add complexity without grounding in encoder decision geometry. Crucially, none provide a self-supervised, geometrically principled definition of recognizability, leaving predictions as proxies rather than encoder-specific signals.

We introduce \textbf{TransFIRA} (Transfer Learning for Face Image Recognizability Assessment), the first framework to adapt any pretrained backbone for recognizability prediction without human quality annotations or IQA supervision. TransFIRA derives supervision directly from embeddings via \emph{class-center similarity} (CCS) and \emph{class-center angular separation} (CCAS), labels deterministically tied to the encoder’s discrimination ability. This yields predictions that are encoder-specific, geometry-driven, and interpretable. Beyond prediction, TransFIRA enables \textbf{recognizability-informed aggregation}: a natural cutoff ($CCAS>0$) that filters unrecognizable frames and CCS-based weighting that emphasizes compact, reliable embeddings. Together, these operations replace heuristics with transparent, geometry-grounded rules.

Extensive experiments on BRIAR Protocol 3.1~\cite{cornett2023expanding} and IJB-C~\cite{maze2018iarpa} show that TransFIRA consistently surpasses prior FIQA methods such as MagFace~\cite{meng2021magface}, CR-FIQA~\cite{Boutros_2023_CVPR}, and DifFIQA~\cite{babnikIJCB2023}, while remaining lightweight and broadly adaptable. Beyond faces, we extend the framework to body recognition via a sigmoid calibration strategy that restores discriminative power for CCS and CCAS. Finally, we show that recognizability predictions provide the first \emph{encoder-grounded explainability}: they capture how degradations such as blur or occlusion alter recognizability, flag subjects who are inherently difficult to identify, and expose conditions where recognition is likely to fail.

\begin{figure*}[t]
  \includegraphics[width=\linewidth]{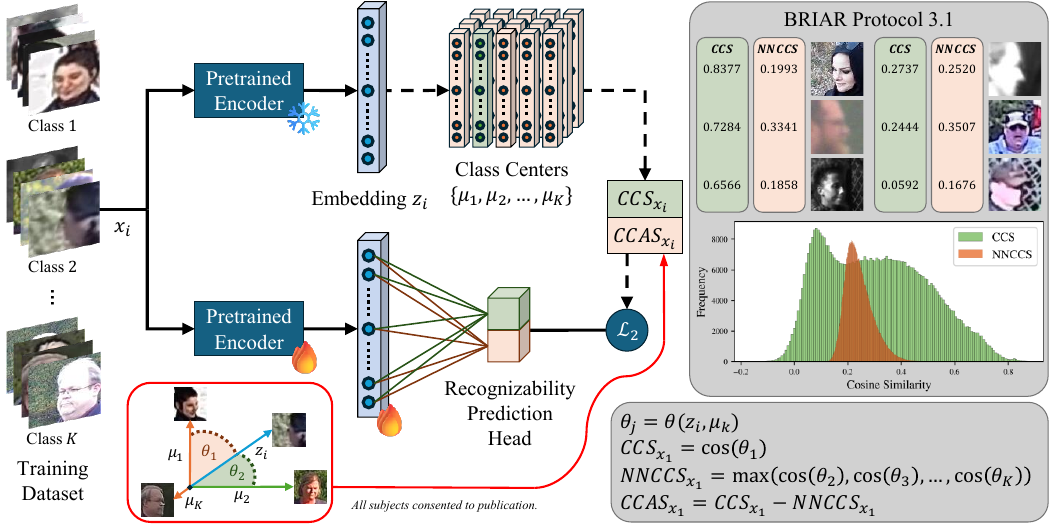}
  \caption{\textbf{Overview of TransFIRA.} Visual quality does not reliably predict recognizability; for instance, blurred faces may still be discriminative while clear ones may fail. TransFIRA avoids such proxies by deriving recognizability labels directly from class-center similarities ($CCS$, $NNCCS$, $CCAS$) and fine-tuning any pretrained encoder end-to-end with a prediction head. Scores remain tied to the encoder’s embedding space, enabling two key operations: \emph{recognizability-weighted aggregation} using $CCS$ and \emph{principled filtering} with the natural cutoff $CCAS>0$. This encoder-agnostic design improves robustness across challenging benchmarks (e.g., BRIAR Protocol 3.1~\cite{cornett2023expanding}) and generalizes naturally to other modalities (Section~\ref{sec:method:body}).}
  \label{fig:teaser}
  \vspace{-3mm}
\end{figure*}

In summary, we propose the following contributions:
\begin{itemize}
\item \textbf{Encoder-specific recognizability supervision:} CCS and CCAS derived directly from embeddings provide the first geometrically grounded basis for FIQA.
\item \textbf{Flexible transfer learning:} Any pretrained encoder can be extended with a lightweight regression head, enabling recognizability prediction without retraining recognition or altering backbone design.
\item \textbf{Recognizability-informed aggregation and explainability:} A natural $CCAS>0$ cutoff and CCS-based weighting yield state-of-the-art results on BRIAR and IJB-C, generalize to body recognition, and enable encoder-grounded explainability.
\end{itemize}

\section{Related Work}

Research in face image quality assessment (FIQA) has evolved along three main directions. Analytical methods estimate quality from hand-crafted image statistics such as sharpness, pose, or illumination~\cite{raghavendra2014automatic, lijun2019multi, gao2007standardization, terhorst2020ser, IconicityFace}. While simple, these approaches generalize poorly because visual degradations do not consistently align with recognizability, leaving their scores detached from the decision geometry of modern recognition models.

Regression and uncertainty-based methods learn mappings from images or embeddings to continuous quality scores. FaceQNet~\cite{hernandez2019faceqnet} fine-tunes a ResNet-50 on automatically labeled VGGFace2 data to approximate commercial system behavior, SDD-FIQA~\cite{SDD-FIQA2021} introduces an unsupervised similarity distribution distance, and SER-FIQ~\cite{DBLP:conf/cvpr/TerhorstKDKK20} estimates reliability via Monte Carlo dropout. While correlated with accuracy, these methods depend on curated labels, auxiliary supervision, or stochastic sampling, making them less direct and more resource-intensive.

Model-based methods integrate FIQA directly into recognition training. MagFace~\cite{meng2021magface} encodes quality through embedding magnitudes, CR-FIQA~\cite{boutros2023cr} learns relative classifiability in angular-margin space, and GraFIQs~\cite{DBLP:conf/cvpr/KolfDB22} derives quality from gradient magnitudes induced by batch-normalization discrepancies. While effective, these approaches are tightly coupled to specific backbones and objectives, limiting their transferability to arbitrary encoders.

Recent work has also explored heavier alternatives. CLIB-FIQA~\cite{Ou_2024_CVPR} fits quality with multi-factor calibration via a vision–language framework, while DifFIQA~\cite{babnikIJCB2023} and eDifFIQA~\cite{babnikTBIOM2024} perturb images with diffusion models to approximate recognizability. Though these pipelines advance the state of the art, they remain indirect proxies for encoder behavior and introduce significant computational overhead compared to lightweight embedding-based approaches.

In contrast, TransFIRA provides the first lightweight, self-supervised framework that is encoder-specific and geometrically grounded. By deriving labels directly from class-center similarity and separation, it avoids heuristic proxies, curated labels, and recognition-specific pipelines. TransFIRA adapts seamlessly to any pretrained encoder with only a regression head, produces interpretable decision-boundary–aligned scores, and uniquely extends to body recognition through calibration, establishing the first unified framework for recognizability-aware modeling across modalities.

\section{METHOD}

\textbf{TransFIRA} adapts a pretrained backbone to predict \emph{recognizability}, defined by embedding-space similarity and separation that indicate whether an input will be correctly recognized. Unlike prior FIQA methods based on visual proxies, TransFIRA ties recognizability directly to the encoder’s decision geometry and uses it for interpretable aggregation.

Our framework has three stages. First, we derive recognizability labels directly from embeddings by computing class-center similarities and separations (Section~\ref{sec:method:recognizability}), requiring no human annotations or IQA supervision. Second, we fine-tune the backbone with a lightweight regression head to predict these labels from raw images (Section~\ref{sec:method:network}), producing encoder-specific recognizability scores. Finally, we use these predictions for \textbf{recognizability-informed aggregation} (Section~\ref{sec:method:aggregation}), combining a natural cutoff ($CCAS>0$) that filters unrecognizable frames with CCS-based weighting that emphasizes compact, reliable embeddings. Together, these components form a principled framework that is encoder-specific, geometrically interpretable, and lightweight.

\subsection{Image Recognizability}
\label{sec:method:recognizability}

To supervise recognizability prediction, we derive labels directly from encoder embeddings by comparing each image to the center of its class. This grounds recognizability in the embedding space of the chosen encoder, ensuring that the scores reflect its actual discrimination ability rather than superficial proxies such as blur or illumination. In this way, recognizability labels are obtained automatically from the encoder itself, without relying on handcrafted measures or other IQA supervision.

Let $\mathcal{D} \,{=}\, \{(x_i,\,y_i)\}_{i=1}^{N}$ be a training dataset of $N$ face images $x_i \in \mathbb{R}^{H \times W \times 3}$ with identity labels $y_i \in \{1,\dots,K\}$, where $K$ is the number of classes. A pretrained face encoder $\phi$ maps each image to a $d$-dimensional embedding:
\setlength{\abovedisplayskip}{5.5pt plus 1pt minus 2pt}
\setlength{\belowdisplayskip}{5.5pt plus 1pt minus 2pt}
\setlength{\abovedisplayshortskip}{3pt plus 1pt}
\setlength{\belowdisplayshortskip}{3pt plus 1pt minus 1pt}
\begin{equation}
z_i = \phi(x_i).
\end{equation}
We measure the \textbf{recognizability} of an image $x_i$ by its similarity to the \textbf{class center} of its identity. When a separate gallery set $\mathcal{G}$ is available, class centers are computed only from gallery embeddings:
\begin{equation}
\mu_j \;=\; \frac{1}{n_j^{\mathcal{G}}} \sum_{\substack{i \in \mathcal{G} \\ y_i = j}} z_i,
\end{equation}
where $n_j^{\mathcal{G}}$ is the number of gallery samples for class $j$. If no gallery–probe split exists, class centers are computed from all embeddings of identity $j$ in $\mathcal{D}$:
\begin{equation}
\mu_j \;=\; \frac{1}{n_j} \sum_{i:\, y_i = j} z_i,
\end{equation}
where $n_j$ is the number of samples for class $j$. 

We define the \textbf{Class Center Angular Similarity (CCS)} as the cosine similarity between $z_i$ and its class center $\mu_{y_i}$:
\begin{equation}
CCS_{x_i} = \frac{z_i^\top \mu_{y_i}}{\|z_i\|_2 \, \|\mu_{y_i}\|_2},
\end{equation}
which captures how closely an embedding aligns with its own class center. 

We then define the \textbf{Nearest Nonmatch Class Center Angular Similarity (NNCCS)} as the maximum similarity to any other class center:
\begin{equation}
NNCCS_{x_i} = \max_{\,j \neq y_i} \frac{z_i^\top \mu_j}{\|z_i\|_2 \, \|\mu_j\|_2},
\end{equation}
capturing how strongly the image resembles its most confusable impostor class. Fig.~\ref{fig:teaser} illustrates how CCS and NNCCS are computed and distributed on BRIAR Protocol~3.1~\cite{cornett2023expanding}, highlighting their role in defining recognizability.  

Finally, we define the \textbf{Class Center Angular Separation (CCAS)} as their difference:
\begin{equation}
CCAS_{x_i} = CCS_{x_i} - NNCCS_{x_i}.
\end{equation}
Large positive values of $CCAS_{x_i}$ indicate strong confidence in the correct class, negative values imply misclassification, and values near zero suggest ambiguity. Importantly, a natural cutoff emerges at $CCAS > 0$: the encoder assigns higher similarity to the correct class center than to any impostor, placing the sample on the correct side of the decision boundary. This provides a principled and interpretable criterion for recognizability, which we use for filtering in Section~\ref{sec:method:aggregation}.

Although CR-FIQA’s \textbf{Certainty Ratio (CR)}~\cite{Boutros_2023_CVPR} is also a function of $CCS$ and $NNCCS$, it does not align with the encoder’s decision geometry. CR is defined as a ratio of match to impostor similarity with an added constant in the denominator, which distorts the underlying margin. For example, consider two samples with identical separations $CCS-NNCCS=0.1$: one with $(CCS,NNCCS)=(0.9,0.8)$ and another with $(0.3,0.2)$. Both are equally well-separated in the embedding space ($CCAS=0.1$), yet their CR values differ sharply ($0.5$ versus $0.25$), leading to inconsistent judgments of recognizability.  

By contrast, both $CCS$ and $CCAS$ map directly to geometric properties of the encoder’s embedding space: $CCS$ reflects alignment with the correct class center, and $CCAS$ reflects separation from impostors. The intrinsic cutoff $CCAS>0$ therefore provides the first principled definition of recognizability: a sample is recognizable precisely when it lies closer to its own class center than to any impostor. This direct interpretability makes CCS and CCAS not only more faithful to the encoder’s decision rule but also more transparent for explainability analyses.

This formulation ties recognizability labels to the encoder’s embedding geometry, ensuring that both labels and thresholds reflect its actual discrimination ability. Because it depends only on embeddings and class centers, it generalizes naturally across modalities. In Section~\ref{sec:method:body}, we extend it to body recognition, where training details and model characteristics cause raw CCS and NNCCS to saturate near one; sigmoid calibration restores variation, yielding well-behaved scores and preserving CCAS as a reliable measure of separability. Appendix~\ref{sec:appendix:explainability} further shows how CCS and CCAS provide encoder-grounded explainability, revealing how degradations like Gaussian blur affect recognizability. Together, CCS, NNCCS, and CCAS offer a geometrically grounded basis for recognizability, independent of handcrafted quality labels or supervision from other IQA methods, and always specific to the encoder in use.

\subsection{Recognizability Prediction Network}
\label{sec:method:network}

Because recognizability labels are deterministic once embeddings are computed, they can be precomputed and reused during training. We predict recognizability directly from images by extending the pretrained encoder $\phi$ with a lightweight \textbf{recognizability prediction head} $h_\psi$, implemented as a linear layer. Given an input $x_i$, the model outputs a two-dimensional recognizability vector:
\begin{equation}
\hat{\mathbf{r}}_i = [\hat{CCS}_{x_i},\, \hat{CCAS}_{x_i}]^\top = h_\psi(\phi(x_i)).
\end{equation}

The network is trained end-to-end, with both the backbone and prediction head optimized jointly, using mean squared error against ground-truth recognizability labels:
\begin{align}
\mathbf{r}_i &\equiv [CCS_{x_i},\, CCAS_{x_i}]^\top, \\
\mathcal{L} &= \frac{1}{N}\sum_{i=1}^N \|\hat{\mathbf{r}}_i - \mathbf{r}_i \|_2^2 .
\end{align}

By explicitly regressing CCS and CCAS, recognizability is tied to the discrimination ability of the \emph{chosen encoder}. The same image may be highly recognizable to one model yet ambiguous to another, depending on their embedding spaces. This encoder-specific property is central: predicted scores reflect the behavior of the deployed system rather than generic image quality. Fine-tuning the entire backbone alongside the prediction head ensures recognizability is learned as an integrated extension of the encoder, rather than as a detached add-on.

As shown in Table~\ref{tab:component_ablation}, full end-to-end training with pretrained initialization provides clear gains over both random initialization and head-only fine-tuning. The models converge after only a single epoch when initialized from a pretrained backbone, compared to roughly 20 epochs without, underscoring the efficiency benefits of transfer learning. The same table also ablates recognizability supervision with CCS, CCAS, and CR-FIQA’s Certainty Ratio (CR)~\cite{Boutros_2023_CVPR}, where CR is computed from CCS and NNCCS. Among these, CCS provides the most stable supervisory signal for training, while CCAS offers a principled margin-based criterion that we later exploit for filtering. By contrast, CR’s ratio form distorts the underlying margin and leads to less reliable supervision, making CCS a more effective alternative.

This embedding-space formulation distinguishes our approach from prior FIQA methods that rely on visual proxies~\cite{hernandezortega2019faceqnetqualityassessmentface,8987255,SDD-FIQA2021,DBLP:conf/cvpr/KolfDB22}, perturbation-based uncertainty~\cite{DBLP:conf/cvpr/TerhorstKDKK20}, or generative/multimodal pretraining~\cite{babnikIJCB2023,babnikTBIOM2024,Ou_2024_CVPR}. Unlike methods that entangle recognizability with recognition~\cite{meng2021magface, Boutros_2023_CVPR}, our framework decouples the two tasks, preserving recognition accuracy while learning recognizability as a separate, embedding-grounded signal. This two-stage design makes the approach encoder-specific, geometrically interpretable, and readily extensible across modalities. Importantly, because recognizability is predicted at the image level, it can also serve as an \emph{explainability signal}: by exposing how the encoder itself perceives reliability under perturbations, our method enables analyses that were previously infeasible. As shown in Appendix~\ref{sec:appendix:explainability}, predicted CCS and CCAS track recognizability dynamics, providing a principled way to identify hard-to-recognize subjects, anticipate failure cases, or guide data augmentation strategies for robustness.

% \begin{table}[t]
% \centering
% \caption{\textbf{Ablation by training scope.} 
% Comparison on BRIAR Protocol~3.1.0 with a Swin Transformer backbone. 
% \emph{Average} denotes uniform template aggregation without recognizability modeling. 
% \emph{Head-only} fine-tunes only the recognizability head on top of the frozen backbone. 
% \emph{No Pretraining} trains the entire network end to end from random initialization. 
% \emph{Backbone + Head} fine-tunes the entire pretrained network end to end. 
% Best metrics are \textbf{bolded}.}
% \begin{tabular}{l|ccc}
% \toprule
% \textbf{Method} & \textbf{TAR@1e-3} & \textbf{TAR@1e-4} & \textbf{TAR@1e-6} \\
% \midrule
% Average (Baseline)            & 0.9077 & 0.9423 & 0.7500 \\
% Head-only    & 0.9231 & 0.8654 & 0.7885 \\
% No Pretraining & 0.9500 & 0.8923 & 0.8154  \\
% % Backbone + Head ($CR$) & 0.9577 & 0.9038 & 0.8538 \\
% Backbone + Head   & \textbf{0.9615} & \textbf{0.9038} & \textbf{0.8462} \\
% \bottomrule
% \end{tabular}
% \label{tab:component_ablation}
% \end{table}

\begin{table}[t]
\centering
\caption{\textbf{Ablation by training scope and predicted label.} 
Evaluation on BRIAR Protocol~3.1~\cite{cornett2023expanding} with CosFace~\cite{wang2018cosface}, reporting Spearman correlation (SC) with ground truth and FNMR–ERC AUC at fixed target FMRs. 
\emph{P} indicates whether the backbone is pretrained, 
\emph{B} and \emph{H} whether the backbone and head are finetuned,
\emph{Score} denotes the label being regressed. 
The highest SC and lowest AUCs are \textbf{bolded}.}
% \resizebox{\columnwidth}{!}{
\begin{tabular}{cccc|cccc}
\toprule
\multicolumn{4}{c|}{\textbf{Method}} & \multicolumn{4}{c}{\textbf{BRIAR~\cite{cornett2023expanding}: FNMR-ERC AUC}}  \\
P & B & H & Label & SC & $10^{-3}$ & $10^{-4}$ & $10^{-6}$ \\
\midrule
\checkmark &  & \checkmark & CCS & 0.4858 & 0.2422 & 0.4020 & 0.6278  \\
 & \checkmark & \checkmark & CCS & 0.7448 & 0.1791 & 0.2990 & 0.4994 \\
\checkmark & \checkmark & \checkmark & CR~\cite{Boutros_2023_CVPR} & 0.8553 & 0.1549 & 0.2739 & 0.4770 \\
\checkmark & \checkmark & \checkmark & CCAS & 0.8230 & 0.1558 & 0.2734 & \textbf{0.4752}  \\
\checkmark & \checkmark & \checkmark & CCS & \textbf{0.8566} & \textbf{0.1536} & \textbf{0.2722} & 0.4753 \\
\bottomrule
\end{tabular}
% }
\label{tab:component_ablation}
\vspace{-6mm}
\end{table}

We refer to this framework as \textbf{TransFIRA (Transfer Learning for Face Image Recognizability Assessment)}. It offers a general-purpose pipeline in which any pretrained backbone, regardless of architecture or training set, can be extended with recognizability prediction through end-to-end fine-tuning. Training reuses the backbone’s existing components (e.g., normalization layers, dropout, augmentation) while adding only a lightweight regression head. Because labels are derived automatically from CCS and CCAS, no human annotation, IQA supervision, or architectural changes are required. The resulting models integrate seamlessly into downstream systems, enabling recognizability-informed aggregation (Section~\ref{sec:method:aggregation}), cross-modality generalization (Section~\ref{sec:method:body}), and encoder-grounded explainability (Appendix~\ref{sec:appendix:explainability}). Taken together, CCS, CCAS, and their predictions establish the first framework that is (i) encoder-specific, (ii) grounded in decision geometry, and (iii) decoupled from recognition itself -- making image-level recognizability predictions a principled foundation for both template aggregation and explainable, more reliable recognition systems.

\subsection{Recognizability-Informed Template Aggregation}
\label{sec:method:aggregation}

\begin{figure*}[t!]
  \centering
  \includegraphics[width=\linewidth]{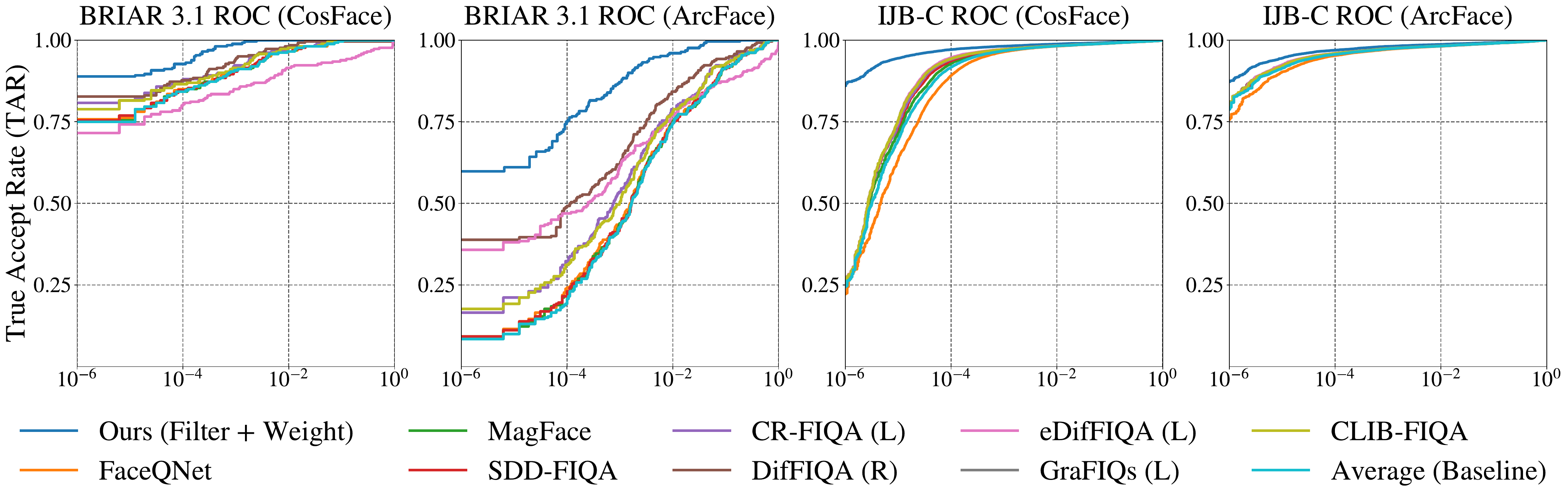}
  % \vspace{-1mm}
  \includegraphics[width=\linewidth]{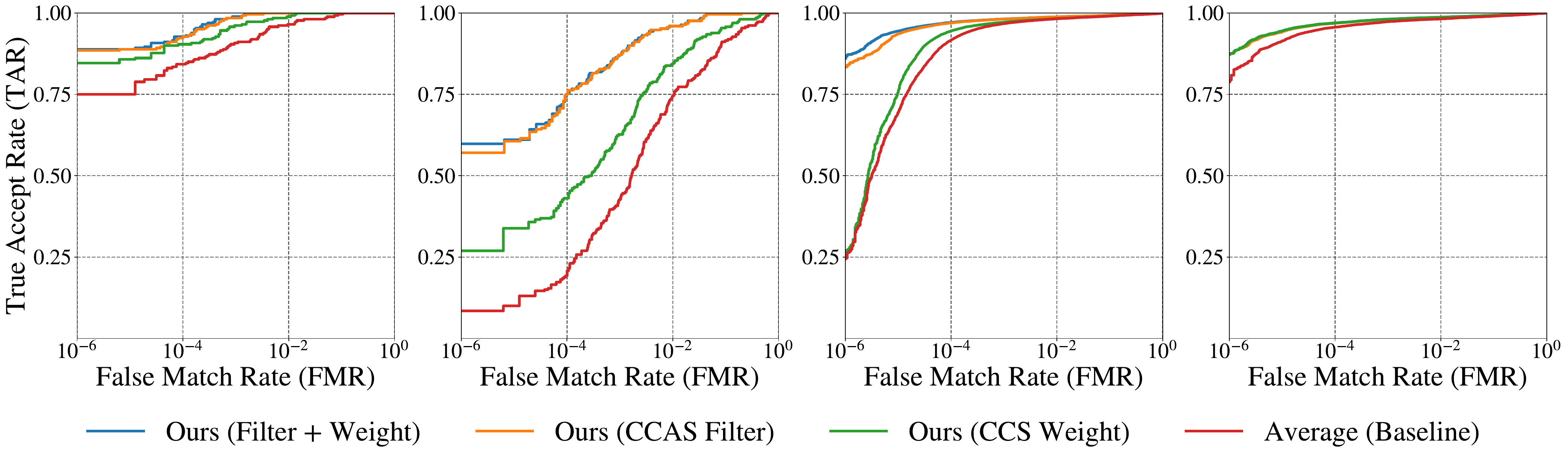}
  \vspace{-6mm}
\caption{\textbf{Overall ROC analysis.} 
Top: template-level ROC comparisons across different IQA methods; for clarity, only the strongest variant of each is shown. 
Bottom: ablation study illustrating the individual and combined effects of CCAS-based filtering and CCS-based weighting. 
Metrics correspond to Table~\ref{tab:tar_results}, and \emph{Average} denotes uniform mean aggregation.}
  \label{fig:roc_all}
  \vspace{-3mm}
\end{figure*}

Image-level recognizability predictions are especially valuable for template aggregation, the standard protocol in BRIAR~\cite{cornett2023expanding}, IJB-C~\cite{maze2018iarpa}, and real-world surveillance, where subjects are represented by many heterogeneous frames. In this setting, even a few unrecognizable images can corrupt the representation and degrade accuracy. A template is typically reduced to a single feature vector by averaging, but uniform weighting treats all frames equally, leaving the system vulnerable to poor-quality inputs. TransFIRA instead exploits recognizability predictions to filter and weight frames, producing cleaner and more discriminative templates.

We first apply \textbf{recognizability-based filtering}, retaining only images with predicted $\hat{CCAS}_{x_i} > 0$. As established in Section~\ref{sec:method:recognizability}, this condition means the sample is predicted to lie on the correct side of the encoder’s decision boundary, so it can be considered recognizable. Unlike tuned thresholds or external quality scores, this natural cutoff is parameter-free, interpretable, and intrinsic to the deployed encoder. We evaluate filtering with CR-FIQA in Appendix~\ref{sec:appendix:filtering_comparison}.

Next, we perform \textbf{recognizability-weighted aggregation}, where each retained embedding is scaled by its predicted $\hat{CCS}_{x_i}$ before averaging:
\begin{equation}
\hat{\mu}^{(s)} =
\frac{\sum_{k \in \mathcal{T}^{(s)}} \hat{CCS}_{x_k}\, z_k}
     {\sum_{k \in \mathcal{T}^{(s)}} \hat{CCS}_{x_k}} ,
\end{equation}
where $\hat{\mu}^{(s)}$ is the aggregated template representation for subject $s$,
and $\mathcal{T}^{(s)}$ denotes the set of images retained after filtering. Because CCS measures proximity to the class center, weighting emphasizes samples that are more compactly embedded within their class, directly reflecting intra-class consistency in the encoder’s embedding space. Rather than relying on visual heuristics or external IQA models, our approach derives weights directly from the deployed encoder itself, ensuring that aggregation is guided by the same geometry in which recognition is performed.

Filtering and weighting are complementary operations that together define a recognizability-informed aggregation strategy -- their individual and combined effects are ablated in Fig.~\ref{fig:roc_all} and Table~\ref{tab:tar_results}. First, filtering introduces the \emph{first principled, parameter-free cutoff}:~$CCAS>0$, which indicates that a frame lies closer to its own class than to any impostor. Second, weighting uses CCS to measure proximity to the class center, favoring embeddings that are more compactly aligned with their class. Unlike prior aggregation schemes that rely on appearance cues or external proxies, both operations are intrinsic to the deployed encoder and map directly onto geometric properties of its embedding space. This grounding makes the method not only effective but also interpretable, as filtering and weighting can be explained transparently in terms of encoder geometry (Appendix~\ref{sec:appendix:explainability}) and extended to other modalities (Section~\ref{sec:method:body}). 

In effect, TransFIRA introduces the first aggregation strategy that is simultaneously principled, encoder-specific, and geometry-driven. It forms a two-layer safeguard for template construction: filtering removes harmful frames, while weighting amplifies the most reliable ones. The result is both stronger recognition performance and clear decision criteria absent from prior aggregation methods.
\begin{table*}[t]
\centering
\caption{\textbf{Overall ROC performance.} TAR at fixed FMRs for BRIAR Protocol~3.1~\cite{cornett2023expanding} and IJB-C~\cite{maze2018iarpa} using the backbones in Section~\ref{sec:experiments:setup}. All baseline methods perform \emph{weighted} aggregation using FIQA scores, while our \textbf{CCAS Filter} and \textbf{Filter + Weight} apply our $CCAS>0$ cutoff. \emph{Average} denotes uniform aggregation without quality weighting. For each column, the best result is \textbf{\underline{bolded and underlined}}, and the second-best is \textbf{bolded}. Appendix~\ref{sec:appendix:filtering_comparison} compares filtering with CR-FIQA~\cite{Boutros_2023_CVPR}.}
\renewcommand{\arraystretch}{1.2}
\resizebox{\textwidth}{!}{%
    \begin{tabular}{l|ccc|ccc|ccc|ccc}
    \toprule
    \multirow{3}{*}{\textbf{Method}} & \multicolumn{6}{c|}{\textbf{BRIAR Protocol 3.1~\cite{cornett2023expanding}: TAR at Fixed FMR}} & \multicolumn{6}{c}{\textbf{IJB-C~\cite{maze2018iarpa}: TAR at Fixed FMR}} \\
    % \cmidrule(lr){2-7} \cmidrule(lr){8-13}
     & \multicolumn{3}{c|}{\textbf{CosFace (BRIAR)~\cite{wang2018cosface, cornett2023expanding}}} & \multicolumn{3}{c|}{\textbf{ArcFace (WebFace)~\cite{guo2021insightface, zhu2021webface260m}}} & \multicolumn{3}{c|}{\textbf{CosFace (BRIAR)~\cite{wang2018cosface, cornett2023expanding}}} & \multicolumn{3}{c}{\textbf{ArcFace (WebFace)~\cite{guo2021insightface, zhu2021webface260m}}} \\
     & $10^{-3}$ & $10^{-4}$ & $10^{-6}$ & $10^{-3}$ & $10^{-4}$ & $10^{-6}$ & $10^{-3}$ & $10^{-4}$ & $10^{-6}$ & $10^{-3}$ & $10^{-4}$ & $10^{-6}$ \\
    \midrule
    Average (Baseline) & 0.9077 & 0.8423 & 0.7500 & 0.4269 & 0.1923 & 0.0846 & 0.9675 & 0.9168 & 0.2440 & 0.9734 & 0.9564 & 0.7881 \\
    FaceQNet~\cite{hernandezortega2021biometricqualityreviewapplication, hernandezortega2019faceqnetqualityassessmentface} & 0.9077 & 0.8500 & 0.7577 & 0.4385 & 0.2385 & 0.0923 & 0.9642 & 0.8912 & 0.2224 & 0.9717 & 0.9529 & 0.7567 \\
    MagFace~\cite{meng2021magface} & 0.9077 & 0.8423 & 0.7538 & 0.4385 & 0.2154 & 0.0923 & 0.9694 & 0.9280 & 0.2504 & 0.9736 & 0.9575 & 0.7970 \\
    SDD-FIQA~\cite{SDD-FIQA2021} & 0.9038 & 0.8462 & 0.7538 & 0.4269 & 0.2154 & 0.0923 & 0.9708 & 0.9361 & 0.2525 & 0.9738 & 0.9590 & 0.7977 \\
    CR-FIQA (S)~\cite{Boutros_2023_CVPR} & 0.9038 & 0.8462 & 0.7577 & 0.4538 & 0.2577 & 0.1077 & 0.9721 & 0.9434 & 0.2522 & 0.9742 & 0.9595 & 0.7966 \\
    CR-FIQA (L)~\cite{Boutros_2023_CVPR} & 0.9231 & 0.8731 & 0.8077 & 0.5346 & 0.3231 & 0.1654 & 0.9800 & 0.9617 & 0.3711 & \textbf{0.9811} & \textbf{\underline{0.9700}} & \textbf{\underline{0.8726}} \\
    DifFIQA (R)\cite{babnikIJCB2023} & 0.9423 & 0.8769 & 0.8269 & 0.6192 & 0.4885 & 0.3885 & 0.9720 & 0.9447 & 0.2499 & 0.9739 & 0.9596 & 0.7972 \\
    % eDifFIQA (T)~\cite{babnikTBIOM2024} & 0.9038 & 0.8615 & 0.7769 & 0.4654 & 0.3000 & 0.1538 & 0.9709 & 0.9384 & 0.2533 & 0.9737 & 0.9594 & 0.7962 \\
    eDifFIQA (S)~\cite{babnikTBIOM2024} & 0.8885 & 0.8462 & 0.7962 & 0.5500 & 0.3731 & 0.2808 & 0.9720 & 0.9454 & 0.2531 & 0.9739 & 0.9597 & 0.7958 \\
    eDifFIQA (M)~\cite{babnikTBIOM2024} & 0.6769 & 0.5923 & 0.4423 & 0.3962 & 0.3000 & 0.2115 & 0.9717 & 0.9482 & 0.2518 & 0.9740 & 0.9597 & 0.7956 \\
    eDifFIQA (L)~\cite{babnikTBIOM2024} & 0.8500 & 0.8000 & 0.7154 & 0.6154 & 0.4692 & 0.3577 & 0.9721 & 0.9468 & 0.2529 & 0.9739 & 0.9599 & 0.7971 \\
    GraFIQs (S)~\cite{DBLP:conf/cvpr/KolfDB22} & 0.9077 & 0.8423 & 0.7500 & 0.4269 & 0.1923 & 0.0846 & 0.9676 & 0.9175 & 0.2441 & 0.9735 & 0.9565 & 0.7884 \\
    GraFIQs (L)~\cite{DBLP:conf/cvpr/KolfDB22} & 0.9077 & 0.8423 & 0.7500 & 0.4269 & 0.1923 & 0.0846 & 0.9675 & 0.9169 & 0.2440 & 0.9734 & 0.9564 & 0.7881 \\
    CLIB-FIQA~\cite{Ou_2024_CVPR} & 0.9154 & 0.8654 & 0.7885 & 0.5038 & 0.3115 & 0.1769 & 0.9714 & 0.9428 & 0.2513 & 0.9740 & 0.9594 & 0.7979 \\
    \midrule
    % Ours (CCAS Weight) & 0.5846 & 0.5654 & 0.5577 & 0.0731 & 0.0385 & 0.0231 & 0.9698 & 0.9393 & 0.2505 & 0.9738 & 0.9585 & 0.7971 \\
    \textbf{Ours (CCAS Filter)} & \textbf{0.9846} & \textbf{0.9231} & \textbf{0.8846} & \textbf{0.8715} & \textbf{0.7510} & \textbf{0.5703} & \textbf{0.9811} & \textbf{0.9697} & \textbf{0.8334} & \textbf{\underline{0.9811}} & 0.9685 & 0.8712 \\
    \textbf{Ours (CCS Weight)} & 0.9615 & 0.9038 & 0.8462 & 0.6269 & 0.4308 & 0.2692 & 0.9718 & 0.9447 & 0.2512 & 0.9737 & 0.9578 & 0.7970 \\
    \textbf{Ours (Filter + Weight)} & \textbf{\underline{0.9885}} & \textbf{\underline{0.9269}} & \textbf{\underline{0.8885}} & \textbf{\underline{0.8715}} & \textbf{\underline{0.7510}} & \textbf{\underline{0.5984}} & \textbf{\underline{0.9815}} & \textbf{\underline{0.9711}} & \textbf{\underline{0.8601}} & 0.9809 & \textbf{0.9693} & \textbf{0.8724} \\
    \bottomrule
    \end{tabular}%
}
\label{tab:tar_results}
\vspace{-3mm}
\end{table*}

\section{EXPERIMENTS}
We evaluate TransFIRA across face and body recognition tasks to answer four key questions: (i) does recognizability-aware aggregation improve template verification accuracy, (ii) do predicted scores align with ground-truth recognizability at the image level, (iii) how well does the framework generalize across training and evaluation domains, and (iv) can it transfer beyond faces to body recognition? Additional results on explainability (Appendix~\ref{sec:appendix:explainability}) and cross-dataset generalization (Appendix~\ref{sec:appendix:crossdataset}) complement the main experiments.

\subsection{Datasets}
\label{sec:experiments:datasets}

We evaluate on two template-based benchmarks. \emph{Templates} are sets of frames from the same subject aggregated into a single representation, and their quality is sensitive to outliers: even one unrecognizable frame can corrupt the aggregate and degrade accuracy. \textbf{IJB-C~\cite{maze2018iarpa}} contains 3,531 subjects drawn from diverse web and video imagery and is the standard benchmark for unconstrained face recognition with strong open-source baselines. \textbf{BRIAR Protocol 3.1~\cite{cornett2023expanding}} contains 615 subjects~\cite{narayan2025petalface} and focuses on video surveillance scenarios with extreme degradations. Because it is private, open-source encoders have not been exposed to its distribution, making it representative of real-world deployment. In this setting, conventional IQA is not informative because most frames are degraded. However, modern encoders still achieve strong results, underscoring the need for encoder-specific FIQA. BRIAR demographics are reported in~\cite{jager2025expandingbriardatasetcomprehensive}.

Traditional datasets such as LFW~\cite{LFWTech} are saturated at near-perfect accuracy, where nearly all images are trivially recognizable and no longer differentiate FIQA methods. In contrast, IJB-C and BRIAR stress-test recognizability prediction in complementary ways: IJB-C through diverse unconstrained imagery, and BRIAR through severe real-world degradations. Together, they move beyond proxy benchmarks to expose the limitations of methods that rely on visual cues or generic training objectives. This dual evaluation positions FIQA not just as an academic exercise but as a foundation for robust recognition in deployment, highlighting the need for encoder-specific approaches like TransFIRA that remain faithful to the decision geometry of the deployed system.

\subsection{Experimental Setup}
\label{sec:experiments:setup}

We evaluate with two complementary backbones. \textbf{CosFace (BRIAR)} is a Swin-B Transformer trained with CosFace on the private BRIAR dataset, achieving state-of-the-art results on BRIAR Protocol 3.1~\cite{cornett2023expanding,wang2018cosface,liu2021swin}. \textbf{ArcFace (WebFace)} is the open-source iResNet50 that achieves the strongest IJB-C performance among the publicly available models in the InsightFace repository~\cite{deng2019arcface,guo2021insightface,zhu2021webface260m,maze2018iarpa}. Each backbone is extended with a linear layer for CCS and CCAS, trained end-to-end with MSE loss under a unified protocol: disjoint splits (a subset of BRIAR for BRIAR evaluations, 10,000 WebFace12M identities for IJB-C), 50 epochs of AdamW ($10^{-6}$ learning rate, $10^{-4}$ weight decay, $64$ batch size), and checkpoint selection by validation Spearman correlation. This setup isolates the impact of recognizability-aware modeling while spanning both a domain-matched SOTA benchmark and a fully reproducible public baseline. BRIAR and WebFace demographic data are reported in \cite{jager2025expandingbriardatasetcomprehensive, zhu2021webface260m}.

\subsection{Recognizability-Aware Template Aggregation}

\begin{figure*}[t!]
  \centering
  \includegraphics[width=\linewidth]{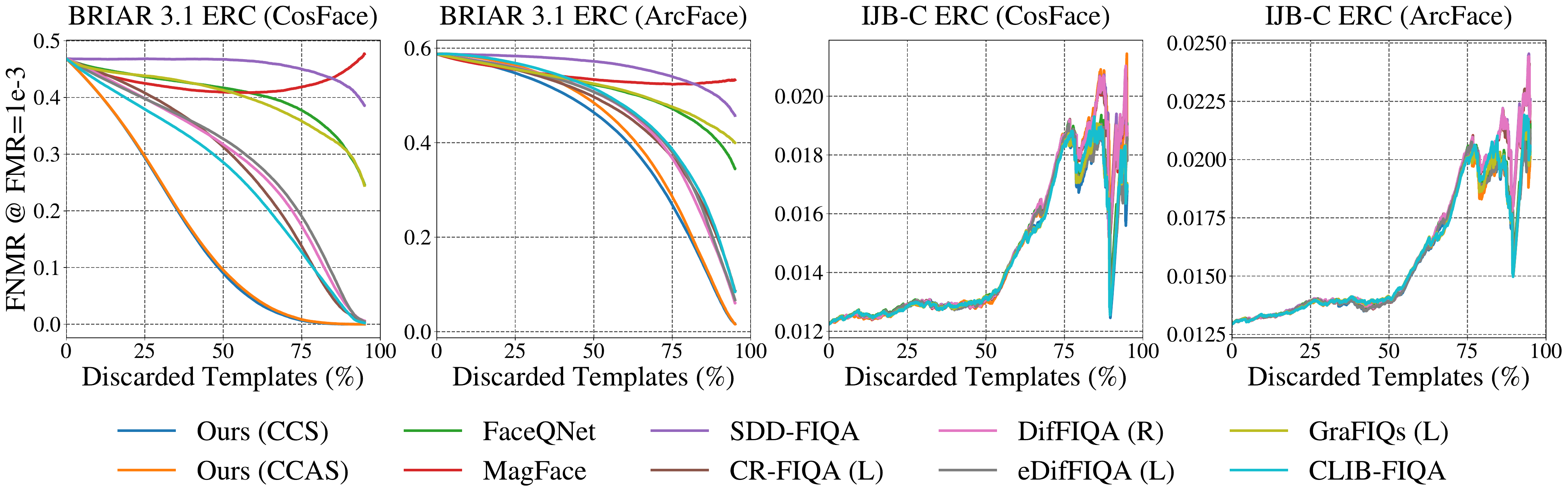}
  \vspace{-7mm}
    \caption{\textbf{Image-level ERC curves at a target FMR of $\mathbf{10^{-3}}$.} 
    Curves closer to the bottom-left indicate better trade-offs between discarding fraction and FNMR. 
    AUCs and Spearman correlations are reported in Table~\ref{tab:erc_auc_results}, with stricter operating points ($10^{-4}$ and $10^{-6}$) reported in Appendix~\ref{sec:appendix:erc}. 
    For clarity, Only the strongest variant of each method is shown.}
  \label{fig:erc_all}
  \vspace{-2mm}
\end{figure*}
\begin{table*}[t]
\centering
\caption{\textbf{Image-level recognizability evaluation.} Image-level Spearman correlation (SC) and FNMR--ERC AUC at fixed FMRs for BRIAR Protocol~3.1~\cite{cornett2023expanding} and IJB-C~\cite{maze2018iarpa}, using the backbones in Section~\ref{sec:experiments:setup}. SC is computed against CCS for all methods except for \textbf{Ours (CCAS)}, where it is computed against CCAS. For each dataset and backbone, the method with the highest SC or lowest AUC is highlighted in \textbf{\underline{bold and underlined}}, while the second-best is shown in \textbf{bold}.
}
\renewcommand{\arraystretch}{1.2}
\resizebox{\textwidth}{!}{%
    \begin{tabular}{l|cccc|cccc|cccc|cccc}
    \toprule
    \multirow{3}{*}{\textbf{Method}} & \multicolumn{8}{c|}{\textbf{BRIAR Protocol 3.1~\cite{cornett2023expanding}: FNMR-ERC AUC}} & \multicolumn{8}{c}{\textbf{IJB-C~\cite{maze2018iarpa}: FNMR-ERC AUC}} \\
    % \cmidrule(lr){2-9} \cmidrule(lr){10-17}
     & \multicolumn{4}{c|}{\textbf{CosFace (BRIAR)~\cite{wang2018cosface, cornett2023expanding}}} & \multicolumn{4}{c|}{\textbf{ArcFace (WebFace)~\cite{guo2021insightface, zhu2021webface260m}}} & \multicolumn{4}{c|}{\textbf{CosFace (BRIAR)~\cite{wang2018cosface, cornett2023expanding}}} & \multicolumn{4}{c}{\textbf{ArcFace (WebFace)~\cite{guo2021insightface, zhu2021webface260m}}} \\
     & SC & $10^{-3}$ & $10^{-4}$ & $10^{-6}$ & SC & $10^{-3}$ & $10^{-4}$ & $10^{-6}$ & SC & $10^{-3}$ & $10^{-4}$ & $10^{-6}$ & SC & $10^{-3}$ & $10^{-4}$ & $10^{-6}$ \\
    \midrule
FaceQNet~\cite{hernandezortega2021biometricqualityreviewapplication, hernandezortega2019faceqnetqualityassessmentface} & 0.1589 & 0.3957 & 0.5225 & 0.6903 & 0.1194 & 0.5037 & 0.7606 & 0.8924 & 0.3607 & 0.0143 & 0.0438 & 0.6716 & 0.3677 & 0.0154 & 0.0288 & 0.1239 \\
    MagFace~\cite{meng2021magface} & 0.1367 & 0.4314 & 0.5616 & 0.7336 & 0.0467 & 0.5426 & 0.8054 & 0.9200 & 0.5684 & 0.0144 & 0.0439 & 0.6663 & 0.6030 & 0.0156 & 0.0290 & 0.1237 \\
    SDD-FIQA~\cite{SDD-FIQA2021} & 0.0293 & 0.4513 & 0.5771 & 0.7335 & 0.0112 & 0.5527 & 0.8069 & 0.9148 & 0.5374 & 0.0144 & 0.0441 & 0.6668 & 0.5682 & 0.0156 & 0.0291 & 0.1242 \\
    CR-FIQA (S)~\cite{Boutros_2023_CVPR} & 0.1107 & 0.4199 & 0.5463 & 0.7093 & 0.0570 & 0.5409 & 0.7874 & 0.8975 & 0.4218 & 0.0142 & 0.0436 & 0.6700 & 0.4589 & 0.0154 & 0.0287 & 0.1232 \\
    CR-FIQA (L)~\cite{Boutros_2023_CVPR} & 0.4143 & 0.2708 & 0.3839 & 0.5640 & 0.3654 & 0.4387 & 0.6815 & 0.8386 & 0.5571 & 0.0144 & 0.0440 & 0.6670 & 0.5902 & 0.0156 & 0.0290 & 0.1237 \\
    DifFIQA (R)~\cite{babnikIJCB2023} & 0.4416 & 0.2782 & 0.3820 & 0.5490 & 0.4657 & 0.4375 & 0.6622 & 0.8194 & 0.4729 & 0.0144 & 0.0440 & 0.6688 & 0.5193 & 0.0156 & 0.0290 & 0.1240 \\
    % eDifFIQA (T) & 0.1736 & 0.3787 & 0.4923 & 0.6510 & 0.1435 & 0.5160 & 0.7522 & 0.8719 & 0.5118 & 0.0145 & 0.0444 & 0.6673 & 0.5508 & 0.0157 & 0.0293 & 0.1253 \\
    eDifFIQA (S)~\cite{babnikTBIOM2024} & 0.3610 & 0.2978 & 0.4064 & 0.5763 & 0.2577 & 0.4644 & 0.7035 & 0.8433 & 0.5236 & 0.0144 & 0.0439 & 0.6672 & 0.5663 & 0.0156 & 0.0290 & 0.1236 \\
    eDifFIQA (M)~\cite{babnikTBIOM2024} & 0.2870 & 0.3249 & 0.4267 & 0.5836 & 0.2265 & 0.4517 & 0.6895 & 0.8316 & 0.5277 & 0.0146 & 0.0444 & 0.6683 & 0.5715 & 0.0158 & 0.0292 & 0.1248 \\
    eDifFIQA (L)~\cite{babnikTBIOM2024} & 0.4377 & 0.2840 & 0.3855 & 0.5489 & 0.4415 & 0.4407 & 0.6658 & 0.8198 & 0.5344 & \textbf{0.0141} & \textbf{\underline{0.0433}} & 0.6674 & 0.5803 & 0.0153 & 0.0286 & \textbf{0.1229} \\
    GraFIQs (S)~\cite{DBLP:conf/cvpr/KolfDB22} & 0.0350 & 0.4363 & 0.5646 & 0.7344 & -0.0194 & 0.5325 & 0.8102 & 0.9287 & 0.1050 & 0.0143 & 0.0438 & 0.6793 & 0.0911 & 0.0154 & 0.0288 & 0.1237 \\
    GraFIQs (L)~\cite{DBLP:conf/cvpr/KolfDB22} & 0.1831 & 0.3881 & 0.5059 & 0.6732 & 0.1434 & 0.5074 & 0.7643 & 0.8914 & 0.3307 & 0.0142 & 0.0436 & 0.6733 & 0.3440 & 0.0154 & 0.0287 & 0.1232 \\
    CLIB-FIQA~\cite{Ou_2024_CVPR} & 0.5004 & 0.2536 & 0.3591 & 0.5340 & 0.4404 & 0.4508 & 0.6758 & 0.8281 & 0.5112 & 0.0142 & 0.0435 & 0.6677 & 0.5580 & 0.0154 & 0.0287 & 0.1231 \\
    \midrule
    \textbf{Ours (CCAS)} & \textbf{0.8230} & \textbf{0.1558} & \textbf{0.2734} & \textbf{\underline{0.4752}} & \textbf{\underline{0.6568}} & \textbf{0.4071} & \textbf{0.6335} & \textbf{\underline{0.8068}} & \textbf{\underline{0.6245}} & 0.0144 & 0.0442 & \textbf{0.6659} & \textbf{\underline{0.6306}} & \textbf{\underline{0.0153}} & \textbf{0.0285} & 0.1230 \\
    \textbf{Ours (CCS)} & \textbf{\underline{0.8566}} & \textbf{\underline{0.1536}} & \textbf{\underline{0.2722}} & \textbf{0.4753} & \textbf{0.6401} & \textbf{\underline{0.3934}} & \textbf{\underline{0.6315}} & \textbf{0.8076} & \textbf{0.6111} & \textbf{\underline{0.0141}} & \textbf{0.0433} & \textbf{\underline{0.6659}} & \textbf{0.6245} & \textbf{0.0153} & \textbf{\underline{0.0285}} & \textbf{\underline{0.1228}} \\
    \bottomrule
    \end{tabular}%
} 
\label{tab:erc_auc_results}
\vspace{-3mm}
\end{table*}

Template aggregation is the primary evaluation protocol in BRIAR and IJB-C, and is critical for real-world applications where recognition must operate over sets of frames rather than single images. Fig.~\ref{fig:roc_all} and Table~\ref{tab:tar_results} evaluate whether recognizability-aware filtering and weighting improve verification accuracy relative to existing FIQA methods.

On BRIAR Protocol 3.1, CCS-weighting achieves the strongest unfiltered performance across operating points, already surpassing prior FIQA methods. Filtering with the intrinsic cutoff $CCAS>0$ provides even larger improvements, and combining filtering with weighting delivers the best results overall. TAR at $10^{-6}$ FMR improves by +0.1385 with Swin and more than triples with ArcFace compared to uniform averaging. To our knowledge, these are the largest FIQA-driven gains yet reported in surveillance settings.

On IJB-C, CCS-weighting remains highly competitive for both backbones, while our CCAS-filtering and combined variants achieve the best results with Swin. For ArcFace, TransFIRA is either the strongest or nearly indistinguishable from the best baseline, trailing CR-FIQA by only 0.001 TAR in some cases. This narrow difference is likely due to training alignment, since both ArcFace and CR-FIQA are optimized on WebFace12M. More broadly, the results suggest that FIQA methods aligned with an encoder’s training distribution can gain slight performance advantages under in-domain evaluation. Appendix~\ref{sec:appendix:filtering_comparison} compares filtering with CR-FIQA.

Overall, recognizability-aware aggregation establishes state-of-the-art TAR while introducing the first explicit geometry-derived filtering rule. On BRIAR, it sets a new frontier for FIQA under surveillance imagery. On IJB-C, it achieves new best results with Swin and nearly matches the strongest ArcFace-aligned baseline. By grounding both filtering and weighting in the encoder’s decision geometry, TransFIRA improves not only accuracy but also interpretability, offering clear and transparent decision rules absent from proxy-based methods. Cross-dataset experiments in Appendix~\ref{sec:appendix:crossdataset} further confirm that these benefits extend beyond the training domain, supporting recognizability as a general principle for robust template aggregation.

\subsection{Image-Level ERC Analysis}
\label{sec:experiments:erc}

While template aggregation measures downstream recognition accuracy, image-level evaluation tests whether predicted recognizability scores faithfully reflect the encoder’s decision geometry. We assess this alignment using two complementary metrics: Spearman correlation (SC), which measures monotonic agreement with ground-truth recognizability, and the FNMR–ERC metric at target FMRs of $10^{-3}$, $10^{-4}$, and $10^{-6}$, which quantifies the impact of recognizability-based filtering. Fig.~\ref{fig:erc_all} and Table~\ref{tab:erc_auc_results} summarize these results, with stronger alignment indicated by higher SC and lower ERC AUC (area under the curve).

On BRIAR Protocol 3.1, TransFIRA substantially outperforms all baselines. With Swin, SC reaches 0.86, nearly double the best competing method (below 0.45), while ERC AUCs are cut almost in half. ArcFace shows the same trend, with both CCS and CCAS providing consistently faithful alignment. These findings confirm that grounding recognizability in encoder geometry captures discriminability far more accurately than visual proxies, directly explaining the large aggregation gains reported above.

On IJB-C, where diffusion-based and multimodal FIQA pipelines compete, TransFIRA again ranks first or second across nearly all metrics and backbones. CCS and CCAS deliver the highest SC and lowest ERC AUCs at most operating points, with a minor exception at $10^{-4}$ for Swin, where we trail eDifFIQA~\cite{babnikTBIOM2024} by 0.001 AUC. Despite its simplicity, TransFIRA matches or surpasses these more complex pipelines while remaining lightweight and encoder-specific.

Overall, these results establish TransFIRA as both the most accurate and the most principled framework for image-level FIQA. By grounding recognizability in CCS and CCAS, it provides the first geometrically interpretable confidence signal without generative modeling or supervision by other IQA methods. This strong image-level fidelity naturally extends to template aggregation, confirming that TransFIRA unites accuracy, interpretability, and efficiency in a way unmatched by prior FIQA approaches. Appendix~\ref{sec:appendix:explainability} further demonstrates how TransFIRA enables encoder-grounded explainability, revealing both failure modes and hard-to-recognize subjects.

\subsection{Body Recognizability}
\label{sec:method:body}

\begin{figure}
    \centering
    \includegraphics[width=\linewidth]{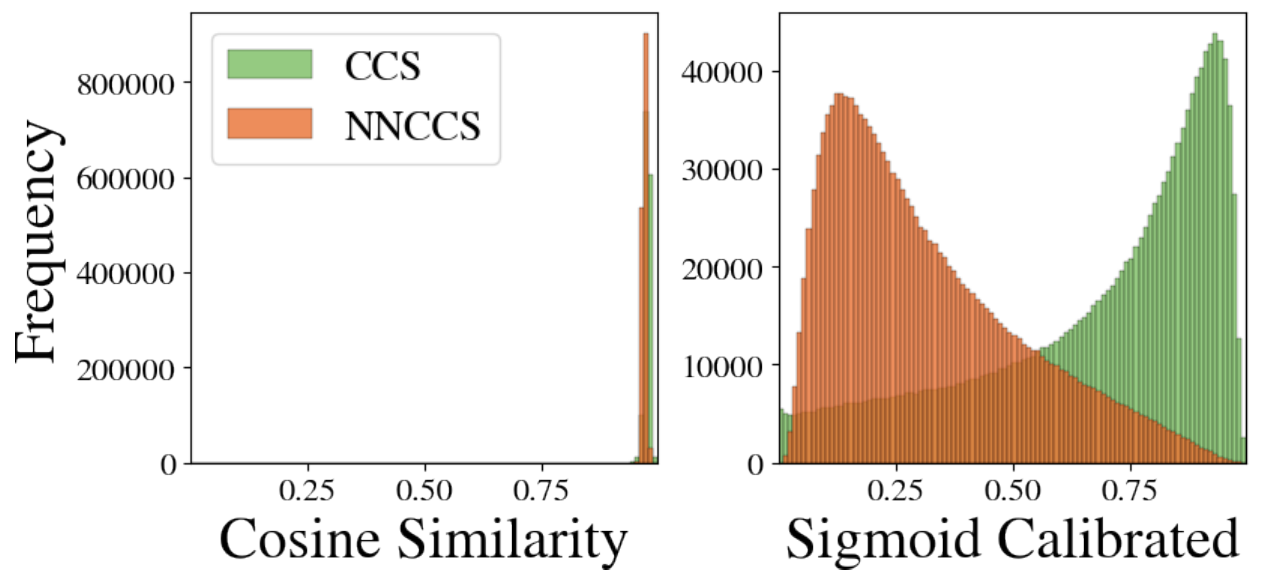}
\caption{\textbf{Sigmoid calibration of BRIAR~\cite{cornett2023expanding} recognizability labels for the SemReID~\cite{zhu2022semreid} body encoder.}
Left: raw CCS/NNCCS distributions (0.97 mean).
Right: sigmoid calibrated distributions (0.50 mean). Appendix~\ref{sec:appendix:body} reports full metrics, including raw and calibrated CCS/CCAS variants.}
\vspace{-3mm}
    \label{fig:sigmoid_calibration}
\end{figure}

To evaluate whether TransFIRA extends beyond faces, we apply it to body recognition on BRIAR Protocol~3.1~\cite{cornett2023expanding} using the SemReID encoder~\cite{zhu2022semreid}. This domain is especially challenging since clothing changes, pose variation, and occlusion often overwhelm appearance-based similarity. As in our face experiments, recognizability labels are derived from cosine similarity to the correct (CCS) and nearest nonmatch (NNCCS) class centers.

Fig.~\ref{fig:sigmoid_calibration} shows that raw CCS and NNCCS values collapse near one, providing little discriminative variation. To address this, we introduce \textbf{sigmoid calibration}, which maps CCS values toward one and NNCCS values toward zero, spreading both across the $[0,1]$ interval. This restores meaningful variation, allowing CCAS to be computed stably and recognizability labels to reflect separability rather than saturation. While sigmoid rescaling is standard in calibration, its application here to recognizability labels is, to our knowledge, novel in the context of FIQA and body recognition.

\begin{table}[t]
\centering
\caption{\textbf{TAR at fixed FMRs for body recognition on BRIAR Protocol~3.1~\cite{cornett2023expanding} using the SemReID encoder~\cite{zhu2022semreid}.} 
The method with the best performance for each operating point is \textbf{bolded}. Full results are reported in Appendix~\ref{sec:appendix:body}.}
\resizebox{\columnwidth}{!}{
\begin{tabular}{l|ccc}
\toprule
\textbf{Method} & \textbf{TAR@1e-3} & \textbf{TAR@1e-4} & \textbf{TAR@1e-6} \\
\midrule
% ----- Test = BRIAR -----
Average (Baseline)  & 0.5934 & 0.3278 & 0.0851 \\
Calibrated CCAS Filter  & 0.6037 &  0.3485 &  0.0933 \\
Calibrated CCAS Weight  & \textbf{0.6452} &  \textbf{0.3693}  & 0.1037 \\
Calibrated Filter + Weight  &  0.6328 &  0.3568  & \textbf{0.1079} \\
\bottomrule
\end{tabular}
}
\label{tab:body}
\vspace{-3mm}
\end{table}

We present the strongest calibrated CCAS results in Table~\ref{tab:body}, where weighting consistently raises TAR the most and combining filtering with weighting provides complementary gains at the strictest operating point. This indicates that \emph{relative separability} (CCAS) is a more reliable indicator of recognizability in body recognition than absolute similarity, which tends to saturate. We therefore adopt CCAS as the primary signal for body recognizability. Appendix~\ref{sec:appendix:body} reports the full results, including raw CCS and CCAS as well as calibrated CCS and CCAS variants, all of which confirm that CCAS outperforms CCS. These findings extend recognizability prediction beyond faces for the first time, establishing TransFIRA as a unified framework for modeling recognizability across modalities.
\section{LIMITATIONS AND FUTURE WORK}

While TransFIRA is compact in that it adds only a lightweight regression head, it still relies on transfer learning with a full backbone, which ties recognizability prediction directly to the pretrained encoder. A natural alternative is to distill recognizability into smaller student models, enabling lighter deployment at the cost of slower convergence and weaker alignment with the encoder’s geometry. Beyond biometrics, encoder-grounded reliability prediction can extend to other domains where decisions depend on embedding quality like reconstruction. Our explainability experiments in Appendix~\ref{sec:appendix:explainability} further suggest that recognizability predictions are not only interpretable but also actionable. They can flag hard-to-identify subjects, reveal operational failure modes such as blur or occlusion, and guide targeted data collection, highlighting their potential as practical tools for system monitoring and robustness.

\section{CONCLUSION}

We introduced \textbf{TransFIRA}, a transfer learning framework that redefines FIQA by grounding recognizability directly in the geometry of the deployed encoder. Unlike prior methods that rely on visual proxies, heuristics, or generative pipelines, TransFIRA derives supervision from embeddings through class-center similarity and angular separation, offering the first self-supervised and geometrically interpretable foundation for recognizability.

Our approach contributes three advances: a principled filtering rule via the natural decision-boundary cutoff ($CCAS>0$), CCS-based weighting for robust template aggregation, and encoder-grounded explainability that exposes how degradations alter recognizability. Experiments on BRIAR\cite{cornett2023expanding} and IJB-C~\cite{maze2018iarpa} show clear state-of-the-art gains, while extensions to body recognition through sigmoid calibration demonstrate robustness across modalities. By closing the gap between visual quality and recognizability, TransFIRA establishes the first annotation-free, geometry-driven framework for recognizability-aware modeling, advancing recognition in both accuracy and interpretability.

\clearpage

\ifFGfinal
\section*{ACKNOWLEDGEMENTS}

This research is based upon work supported in part by the Office of the Director of National Intelligence (ODNI), Intelligence Advanced Research Projects Activity (IARPA), via [2022-21102100005]. The views and conclusions contained herein are those of the authors and should not be interpreted as necessarily representing the official policies, either expressed or implied, of ODNI, IARPA, or the U.S. Government. The US Government is authorized to reproduce and distribute reprints for governmental purposes notwithstanding any copyright annotation therein.
\section*{ETHICAL IMPACT STATEMENT}

This research was conducted in accordance with the FG 2026 Ethical Impact Guidelines. We train our models on the BRIAR dataset~\cite{cornett2023expanding} and WebFace12M~\cite{zhu2021webface260m}, and evaluate on both the BRIAR Protocol 3.1 benchmark and the IJB-C benchmark~\cite{maze2018iarpa}. The BRIAR dataset, collected by IARPA under Institutional Review Board (IRB) approval, consists of surveillance imagery acquired with informed oversight and does not pose harm to the individuals represented. All BRIAR subjects whose images appear in this paper explicitly consented to publication. WebFace12M was collected from publicly available internet sources; while individual subject consent is not feasible at this scale, the dataset is broadly used in academic research and released for non-commercial research purposes. The IJB-C dataset contains unconstrained face images and is publicly available; our use complies with its licensing and ethical use standards.

No new participants were recruited for this research, and thus considerations of participant compensation and vulnerable populations are not applicable. All datasets used reflect a broad demographic spectrum without targeted oversampling.

We recognize that face recognition research carries potential risks, particularly regarding privacy, surveillance misuse, and social bias. To mitigate these risks, we restricted our study to IRB-approved and publicly available benchmarks, maintained data security, and limited analysis to aggregate recognition performance. We acknowledge that recognition technologies can have dual-use applications and emphasize that our contributions are intended solely for academic research. Any code or models released will follow community standards for reproducibility while respecting dataset licensing restrictions. The benefits of this work -- advancing recognizability-aware modeling and developing methods that can make recognition systems more transparent and explainable -- outweigh minimal risks. We further advocate for ethical oversight and clear communication of limitations as key strategies to prevent misuse.

\else

\section*{ETHICAL IMPACT CHECKLIST}
\begin{enumerate}
    \item Yes, we read the Ethical Guidelines document. 
    \item Yes, we adhered to the guidelines for Ethical Review Boards. 
    \item Yes, we adhered to the guidelines for Potential Harms to Human Subjects. 
    \item Yes, we adhered to the guidelines for Potential Negative Societal Impacts. 
    \item Yes, we adhered to the guidelines for Risk-Mitigation Strategies. 
\end{enumerate}
\fi

{\small
\bibliographystyle{ieee}
\bibliography{egbib}
}

\clearpage
\appendix
\section{APPENDIX}
This appendix presents additional analyses, including encoder-grounded explainability under perturbations, extended body recognition experiments, cross-dataset generalization, CR-FIQA filtering, and supplementary ERC curves.

\subsection{Encoder-Grounded Explainability}
\label{sec:appendix:explainability}

\begin{figure*}[h]
    \centering
    \includegraphics[width=\linewidth]{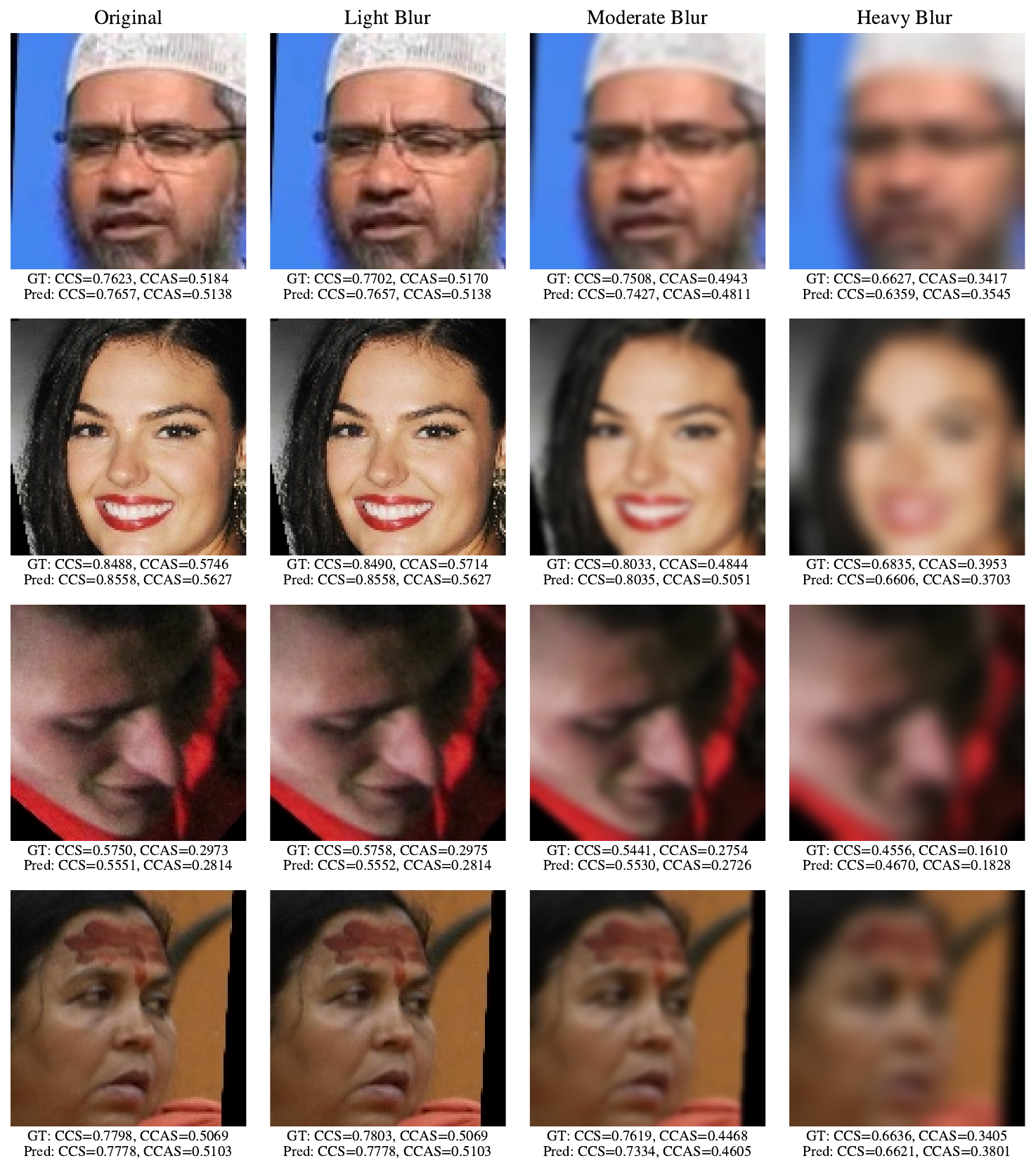}
    \caption{\textbf{Explainability via Gaussian blur with ArcFace~\cite{guo2021insightface,zhu2021webface260m} on IJB-C~\cite{maze2018iarpa}} 
    Across all four examples, adding \emph{light blur} slightly \textbf{improves} recognizability, as both CCS and CCAS increase relative to the no-blur condition. In contrast, \emph{moderate and heavy blur} produce a monotonic decline in recognizability, with the sharpest drop under heavy blur. These patterns highlight that visual degradation does not map directly to recognizability: mild blur can reduce misalignment or suppress distractors, enhancing embeddings, while stronger blur consistently erodes discriminability. This illustrates the value of encoder-grounded signals for capturing nuanced, non-monotonic effects of perturbations.}

    \label{fig:explainability}
\end{figure*}

A distinctive advantage of TransFIRA is that it provides the first \emph{encoder-specific explainability signal} for FIQA. By deriving recognizability directly from embedding geometry, it reveals how the deployed model separates identities rather than relying on visual proxies such as blur, pose, or handcrafted uncertainty. This stands in sharp contrast to prior approaches, which approximate failure modes but remain detached from the encoder’s decision boundary.

We illustrate this with Gaussian blur on IJB-C~\cite{maze2018iarpa}, using an ArcFace encoder finetuned on WebFace12M~\cite{guo2021insightface,zhu2021webface260m}. Blur generally reduces discriminability, but its effect on recognizability is not monotonic. Fig.~\ref{fig:explainability} shows cases where predictions decline smoothly, remain stable, or even improve under mild blur -- when blur suppresses distractors or mitigates misalignment. These examples highlight that visual degradation does not reliably indicate recognizability, underscoring the need for encoder-grounded measures.

Table~\ref{tab:explainability} provides quantitative evidence. While mean CCS and CCAS decrease with stronger blur, predicted values remain strongly correlated with ground truth across all conditions. Spearman correlations reach $\approx$0.62 for CCS and $\approx$0.66 for CCAS without blur and remain high under light and moderate blur, showing that predictions preserve the \emph{relative ordering of recognizability across samples}. This property is critical for explainability: it means predictions not only capture global degradation trends but also identify which images remain robust or degrade most sharply.

A major deployment advantage is that TransFIRA extends to \emph{unseen subjects}, including those with only a single enrollment image where ground-truth CCS/CCAS cannot be computed. In such cases, predicted recognizability is the only practical way to assess whether a subject is reliably identifiable, enabling operators to flag inherently hard-to-recognize individuals, prompt re-enrollment, or add safeguards against likely failures.

High correlation ensures that recognizability predictions reflect the encoder’s own reliability rather than detached quality proxies, making explainability both scalable and actionable. Beyond subject-level monitoring, recognizability dynamics under perturbations can guide augmentation strategies for adversarial robustness and reveal operational conditions most prone to error.

Together, these findings establish TransFIRA as the first framework to provide \emph{encoder-grounded explainability}. By aligning predictions with decision geometry, extending them to unseen subjects, and preserving relative structure under perturbations, TransFIRA yields explanations that are faithful, interpretable, and practically useful -- capabilities absent from prior FIQA methods.

\begin{table}[h]
    \centering
    \caption{\textbf{Effect of Gaussian blur on recognizability.} 
    Mean ground-truth (GT) and predicted (Pred) CCS and CCAS values after applying Gaussian blur of varying intensities to all images in IJB-C, along with their Spearman correlation (SC) across images. All predictions are obtained using an ArcFace encoder finetuned on WebFace12M~\cite{guo2021insightface, zhu2021webface260m}.}

    \resizebox{\columnwidth}{!}{
    \begin{tabular}{l|ccc|ccc}
    \toprule
    \multirow{2}{*}{\textbf{Blur Level}} & \multicolumn{3}{c|}{\textbf{CCS}} & \multicolumn{3}{c}{\textbf{CCAS}} \\
     & GT & Pred & SC & GT & Pred & SC \\
    \midrule
    None     & 0.6335 & 0.7436 & 0.6171 & 0.3351 & 0.4677 & 0.6594 \\
    Light    & 0.6338 & 0.7434 & 0.6157 & 0.3344 & 0.4676 & 0.6596 \\
    Moderate & 0.6040 & 0.7050 & 0.5554 & 0.2891 & 0.4277 & 0.6314 \\
    Heavy    & 0.4865 & 0.5316 & 0.3143 & 0.0834 & 0.2545 & 0.3962 \\
    \bottomrule
    \end{tabular}}
    \label{tab:explainability}
\end{table}

\begin{table}[h!]
\centering
\caption{\textbf{TAR at fixed FMRs for body recognition on BRIAR Protocol~3.1~\cite{cornett2023expanding} with the SemReID encoder~\cite{zhu2022semreid}.} 
For each operating point, the best result is \textbf{\underline{bolded and underlined}}, and the second-best is \textbf{bolded}.}
\resizebox{\columnwidth}{!}{
\begin{tabular}{l|ccc}
\toprule
\textbf{Method} & \textbf{TAR@1e-3} & \textbf{TAR@1e-4} & \textbf{TAR@1e-6} \\
\midrule
% ----- Test = BRIAR -----
Average (Baseline)  & 0.5934 & 0.3278 & 0.0851 \\
\midrule
Raw CCAS Filter  & 0.5934 &  0.3278 &  0.0851 \\
Raw CCS Weight  & 0.5934 &  0.3257  & 0.0851 \\
\hspace{1em} Filter + Weight  &  0.5934 &  0.3257 &  0.0851 \\

Raw CCAS Weight  & 0.6000  & 0.3340  & 0.0851 \\
\hspace{1em} Filter + Weight  & 0.6000 &  0.3340 &  0.0851 \\
\midrule
Calibrated CCAS Filter  & 0.6037 &  0.3485 &  0.0933 \\
Calibrated CCS Weight & 0.6120 &  0.3402  & 0.0934 \\
\hspace{1em} Filter + Weight  &  0.6203  & 0.3465 &  0.0996\\

Calibrated CCAS Weight  & \textbf{\underline{0.6452}} &  \textbf{\underline{0.3693}}  & \textbf{0.1037} \\
\hspace{1em} Filter + Weight  &  \textbf{0.6328} &  \textbf{0.3568}  & \textbf{\underline{0.1079}} \\
\bottomrule
\end{tabular}
}
\label{tab:body_app}
\end{table}
\subsection{Extended Body Recognition Results}
\label{sec:appendix:body}
\begin{table*}[t]
\centering
\caption{\textbf{Cross-dataset generalization.} TAR at fixed FMRs when recognizability prediction is finetuned on one dataset and evaluated on another. 
\textbf{Backbone} indicates the base encoder used in TransFIRA, \textbf{Finetune} the dataset used for recognizability training, \textbf{Encoder} the recognition model being tested, and \textbf{Evaluate} the benchmark dataset. 
Rows marked with * denote domain-matched settings where finetune and evaluation datasets align. 
The best TAR for each evaluation configuration is \textbf{bolded}.}
\resizebox{\linewidth}{!}{
\begin{tabular}{ll|ll|ccc}
\toprule
\textbf{Backbone} & \textbf{Finetune} & \textbf{Encoder} & \textbf{Evaluate} & \textbf{TAR@1e-3} & \textbf{TAR@1e-4} & \textbf{TAR@1e-6} \\
\midrule
% ----- Test = BRIAR~\cite{cornett2023expanding} -----
CosFace~\cite{wang2018cosface}*     & BRIAR~\cite{cornett2023expanding}   & CosFace~\cite{wang2018cosface}     & BRIAR~\cite{cornett2023expanding} & \textbf{0.9615} & \textbf{0.9038} & \textbf{0.8462} \\
ArcFace~\cite{guo2021insightface}  & BRIAR~\cite{cornett2023expanding}   & CosFace~\cite{wang2018cosface}     & BRIAR~\cite{cornett2023expanding} & 0.9423     & 0.8923     & 0.8385     \\
CosFace~\cite{wang2018cosface}     & WebFace~\cite{zhu2021webface260m} & CosFace~\cite{wang2018cosface}     & BRIAR~\cite{cornett2023expanding} & 0.9423     & 0.8923     & 0.8192     \\
ArcFace~\cite{guo2021insightface}  & WebFace~\cite{zhu2021webface260m} & CosFace~\cite{wang2018cosface}     & BRIAR~\cite{cornett2023expanding} & 0.9000     & 0.8423   & 0.7500     \\
\midrule
CosFace~\cite{wang2018cosface}     & BRIAR~\cite{cornett2023expanding}   & ArcFace~\cite{guo2021insightface}  & BRIAR~\cite{cornett2023expanding} & 0.5846 & 0.3538 & 0.2115     \\
ArcFace~\cite{guo2021insightface}*  & BRIAR~\cite{cornett2023expanding}   & ArcFace~\cite{guo2021insightface}  & BRIAR~\cite{cornett2023expanding} & \textbf{0.6269} & \textbf{0.4308} & \textbf{0.2692} \\
CosFace~\cite{wang2018cosface}     & WebFace~\cite{zhu2021webface260m} & ArcFace~\cite{guo2021insightface}  & BRIAR~\cite{cornett2023expanding} & 0.5654 & 0.3423 & 0.1962     \\
ArcFace~\cite{guo2021insightface}  & WebFace~\cite{zhu2021webface260m} & ArcFace~\cite{guo2021insightface}  & BRIAR~\cite{cornett2023expanding} & 0.4154     & 0.2154     & 0.0855     \\
\midrule
% ----- Test = IJB-C~\cite{maze2018iarpa} -----
CosFace~\cite{wang2018cosface}     & BRIAR~\cite{cornett2023expanding}   & CosFace~\cite{wang2018cosface}     & IJB-C~\cite{maze2018iarpa} & \textbf{0.9720} & \textbf{0.9470} & 0.2493     \\
ArcFace~\cite{guo2021insightface}  & BRIAR~\cite{cornett2023expanding}   & CosFace~\cite{wang2018cosface}     & IJB-C~\cite{maze2018iarpa} & 0.9716 & 0.9452 & 0.2255     \\
CosFace~\cite{wang2018cosface}*     & WebFace~\cite{zhu2021webface260m} & CosFace~\cite{wang2018cosface}     & IJB-C~\cite{maze2018iarpa} & 0.9718 & 0.9447 & \textbf{0.2512} \\
ArcFace~\cite{guo2021insightface}  & WebFace~\cite{zhu2021webface260m} & CosFace~\cite{wang2018cosface}     & IJB-C~\cite{maze2018iarpa} & 0.9697 & 0.9276 & 0.2502     \\
\midrule
CosFace~\cite{wang2018cosface}     & BRIAR~\cite{cornett2023expanding}   & ArcFace~\cite{guo2021insightface}  & IJB-C~\cite{maze2018iarpa} & 0.9737 & 0.9594 & 0.7970     \\
ArcFace~\cite{guo2021insightface}  & BRIAR~\cite{cornett2023expanding}   & ArcFace~\cite{guo2021insightface}  & IJB-C~\cite{maze2018iarpa} & 0.9735 & 0.9593 & 0.7958     \\
CosFace~\cite{wang2018cosface}     & WebFace~\cite{zhu2021webface260m} & ArcFace~\cite{guo2021insightface}  & IJB-C~\cite{maze2018iarpa} &  0.9738 & 0.9589 & 0.7978 \\
ArcFace~\cite{guo2021insightface}*  & WebFace~\cite{zhu2021webface260m} & ArcFace~\cite{guo2021insightface}  & IJB-C~\cite{maze2018iarpa} & \textbf{0.9809} & \textbf{0.9693} & \textbf{0.8724} \\
\bottomrule
\end{tabular}
}
\label{tab:cross-dataset}
\end{table*}

Table~\ref{tab:body} reports the full body recognition results on BRIAR Protocol3.1 with the SemReID encoder\cite{zhu2022semreid}, complementing the main findings in Section~\ref{sec:method:body}. As shown in Fig.~\ref{fig:sigmoid_calibration}, raw CCS and NNCCS values collapse near one (mean $0.97$, variance $3.69 \times 10^{-5}$), providing little discriminative variation. Sigmoid calibration spreads the distributions across $[0,1]$ (mean $0.50$, variance $0.09$, Brier score $0.1564$), restoring usable variation and stabilizing CCAS computation.

The extended results confirm that raw CCS- and CCAS-based weighting do not substantially alter performance relative to unweighted aggregation. In contrast, calibration makes a clear difference: calibrated CCAS weighting consistently yields the strongest results across all operating points, outperforming both raw and calibrated CCS. For instance, at $10^{-3}$ and $10^{-4}$ FMR, calibrated CCAS achieves the highest TARs, and at the strictest $10^{-6}$ FMR, filtering combined with calibrated CCAS weighting delivers the best overall performance. These trends demonstrate that calibration is essential when similarity scores collapse into a narrow range and lose discriminative power, with sigmoid calibration restoring discriminative variation and enabling more effective downstream use.

Taken together, these findings establish CCAS, particularly when calibrated, as the most reliable basis for recognizability prediction under conditions of degenerate similarity distributions. In contrast, CCS remains effective for face recognition, where similarity values are less saturated. To our knowledge, this is the first systematic study of recognizability prediction for body recognition, and the results demonstrate that encoder-grounded recognizability is modality-agnostic: while CCS dominates in face recognition, CCAS emerges as the key signal for bodies. This cross-modality flexibility highlights TransFIRA’s adaptability and underscores its practical value for real-world deployments, where recognition systems must operate seamlessly across both facial and bodily inputs.

\subsection{Cross-Dataset Generalization}
\label{sec:appendix:crossdataset}

While domain-matched finetuning yields the strongest results (e.g., CosFace on BRIAR~\cite{wang2018cosface, cornett2023expanding}, ArcFace on WebFace/IJB-C~\cite{deng2019arcface,zhu2021webface260m,maze2018iarpa}), Table~\ref{tab:cross-dataset} shows that TransFIRA remains highly competitive even when training and evaluation domains differ. In several cases, cross-dataset models approach -- or even surpass -- baselines trained in-domain, underscoring that recognizability prediction generalizes more robustly than heuristic or proxy-based FIQA measures.

A particularly notable case is CosFace finetuned on BRIAR but evaluated on IJB-C, which achieves the best TAR at $10^{-3}$ and $10^{-4}$ despite the domain mismatch. This transferability arises from two factors: (i) recognizability supervision is defined by gallery–probe separation, which mirrors IJB-C’s verification protocol, and (ii) CCAS emphasizes relative separability over absolute similarity, making it inherently more resilient to shifts in data distribution. These properties allow TransFIRA to retain discriminative power even when the training and evaluation domains differ substantially.

Cross-dataset generalization is especially challenging in FIQA because datasets differ not only in image quality but also in domain characteristics: BRIAR emphasizes surveillance imagery with severe degradations, while IJB-C emphasizes unconstrained but curated web and video imagery. Despite these differences, TransFIRA consistently aligns with recognition performance across domains. This robustness highlights that recognizability prediction reflects a transferable principle rooted in encoder geometry. While domain-specific finetuning provides the strongest results, the ability to generalize across datasets without retraining is crucial for deployment, where FIQA labels are rarely available for each new distribution and systems must remain reliable under shifting conditions.

\begin{table*}[t]
\centering
\caption{\textbf{CR-FIQA vs.\ Ours: ROC performance with filtering.} 
TAR at fixed FMRs for BRIAR Protocol~3.1~\cite{cornett2023expanding} and IJB-C~\cite{maze2018iarpa} using the backbones in Section~\ref{sec:experiments:setup}.
We directly compare CR-FIQA~\cite{Boutros_2023_CVPR} with a filtering threshold of \textbf{$CR>0.5$} against our encoder-grounded \textbf{$CCAS>0$} rule. 
The best result for each column is highlighted in \textbf{\underline{bold and underlined}}, while the second-best is highlighted in \textbf{bold}. 
Rows labeled ``Filter'' (without $+$) denote filtering alone without weighting.}
\renewcommand{\arraystretch}{1.2}
\resizebox{\textwidth}{!}{%
    \begin{tabular}{l|ccc|ccc|ccc|ccc}
    \toprule
    \multirow{3}{*}{\textbf{Method}} & \multicolumn{6}{c|}{\textbf{BRIAR Protocol 3.1~\cite{cornett2023expanding}: TAR at Fixed FMR}} & \multicolumn{6}{c}{\textbf{IJB-C~\cite{maze2018iarpa}: TAR at Fixed FMR}} \\
    % \cmidrule(lr){2-7} \cmidrule(lr){8-13}
     & \multicolumn{3}{c|}{\textbf{CosFace (BRIAR)~\cite{wang2018cosface, cornett2023expanding}}} & \multicolumn{3}{c|}{\textbf{ArcFace (WebFace)~\cite{guo2021insightface, zhu2021webface260m}}} & \multicolumn{3}{c|}{\textbf{CosFace (BRIAR)~\cite{wang2018cosface, cornett2023expanding}}} & \multicolumn{3}{c}{\textbf{ArcFace (WebFace)~\cite{guo2021insightface, zhu2021webface260m}}} \\
     & $10^{-3}$ & $10^{-4}$ & $10^{-6}$ & $10^{-3}$ & $10^{-4}$ & $10^{-6}$ & $10^{-3}$ & $10^{-4}$ & $10^{-6}$ & $10^{-3}$ & $10^{-4}$ & $10^{-6}$ \\
    \midrule
    CR-FIQA (L)~\cite{Boutros_2023_CVPR} (CR Filter) & 0.9077 & 0.8654 & 0.7808 & 0.4731 & 0.2577 & 0.1000 & 0.9782 & 0.9496 & 0.3617 & 0.9811 & 0.9686 & 0.8711 \\
    CR-FIQA (L)~\cite{Boutros_2023_CVPR} (CR Weight) & 0.9231 & 0.8731 & 0.8077 & 0.5346 & 0.3231 & 0.1654 & 0.9800 & 0.9617 & 0.3711 & 0.9811 & \textbf{0.9700} & \textbf{0.8726} \\
    CR-FIQA (L)~\cite{Boutros_2023_CVPR} (Filter + Weight) & 0.9231 & 0.8808 & 0.8231 & 0.5577 & 0.3385 & 0.1846 & 0.9800 & 0.9620 & 0.3712 & \textbf{0.9811} & \textbf{\underline{0.9701}} & \textbf{\underline{0.8726}} \\
    Ours (CCAS Filter) & \textbf{0.9846} & \textbf{0.9231} & \textbf{0.8846} & \textbf{0.8715} & \textbf{0.7510} & \textbf{0.5703} & \textbf{0.9811} & \textbf{0.9697} & \textbf{0.8334} & \textbf{\underline{0.9811}} & 0.9685 & 0.8712 \\
    Ours (CCS Weight) & 0.9615 & 0.9038 & 0.8462 & 0.6269 & 0.4308 & 0.2692 & 0.9718 & 0.9447 & 0.2512 & 0.9737 & 0.9578 & 0.7970 \\
    Ours (Filter + Weight) & \textbf{\underline{0.9885}} & \textbf{\underline{0.9269}} & \textbf{\underline{0.8885}} & \textbf{\underline{0.8715}} & \textbf{\underline{0.7510}} & \textbf{\underline{0.5984}} & \textbf{\underline{0.9815}} & \textbf{\underline{0.9711}} & \textbf{\underline{0.8601}} & 0.9809 & 0.9693 & 0.8724 \\
    \bottomrule
    \end{tabular}%
}
\label{tab:filtering_comparison}
\end{table*}

\subsection{CR-FIQA Filtering}
\label{sec:appendix:filtering_comparison}
CR-FIQA~\cite{Boutros_2023_CVPR} introduces the Certainty Ratio (CR) as a confidence measure but does not specify a filtering rule. For completeness, we evaluate a natural candidate: $CR>0.5$, the midpoint of the $[0,1]$ scale covered by CR. From the definition,
\[
CR = \frac{CCS}{NNCCS + 1 + \epsilon},
\]
where $\epsilon=10^{-9}$ is a numerical stabilizer, this cutoff effectively requires the genuine similarity to exceed half of the shifted nonmatch similarity. However, the addition of $1+\epsilon$ in the denominator arbitrarily shifts the CR scale, meaning the $0.5$ threshold has no intrinsic connection to the decision geometry. In contrast, our analytical $CCAS>0$ rule corresponds exactly to the condition $CCS>NNCCS$ that defines correct recognition. To our knowledge, this work provides the first systematic evaluation of CR-FIQA as a filtering rule.

As shown in Table~\ref{tab:filtering_comparison}, the difference is consequential. Filtering with $CR>0.5$ adds little beyond weighting alone, whereas $CCAS>0$ filtering improves TAR even on its own and, when combined with weighting, consistently outperforms $CR>0.5$ across both Swin and ArcFace backbones on BRIAR and IJB-C. For instance, with Swin on BRIAR, $CCAS>0$ filtering achieves $0.9846$ TAR at $10^{-3}$ FMR compared to $0.9077$ for $CR>0.5$, and the combined Filter+Weight variant further boosts performance to $0.9885$. These results highlight that while CR-FIQA’s weighting remains strong, its implicit midpoint filtering is arbitrary and of limited value, whereas our decision-boundary–grounded $CCAS>0$ criterion offers both interpretability and measurable gains.

\subsection{Additional Image-Level ERC Curves}
\label{sec:appendix:erc}

\begin{figure*}[t]
  \centering
  \includegraphics[width=\linewidth]{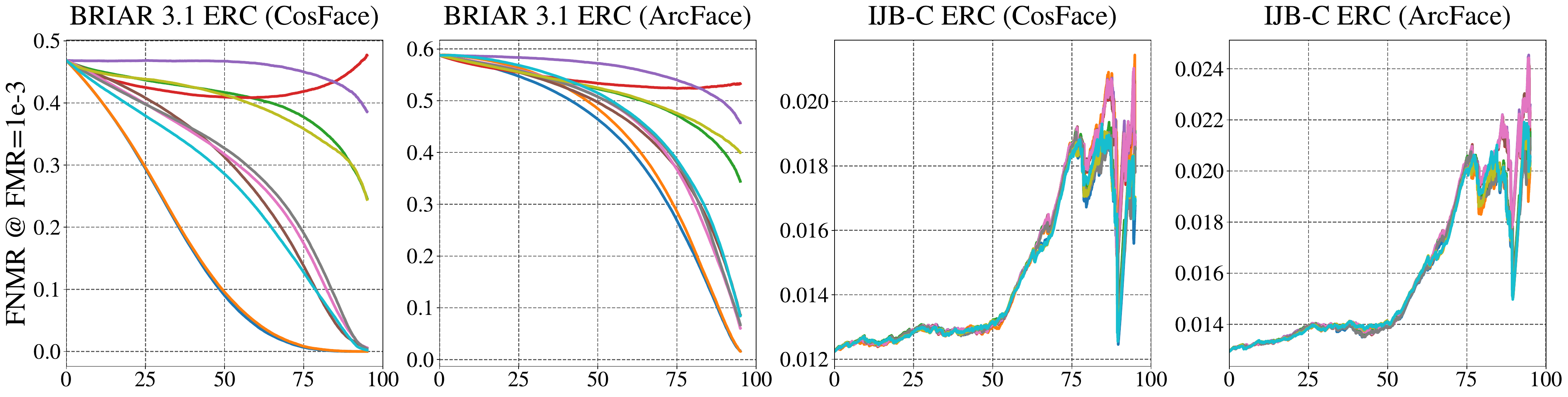}
  \includegraphics[width=\linewidth]{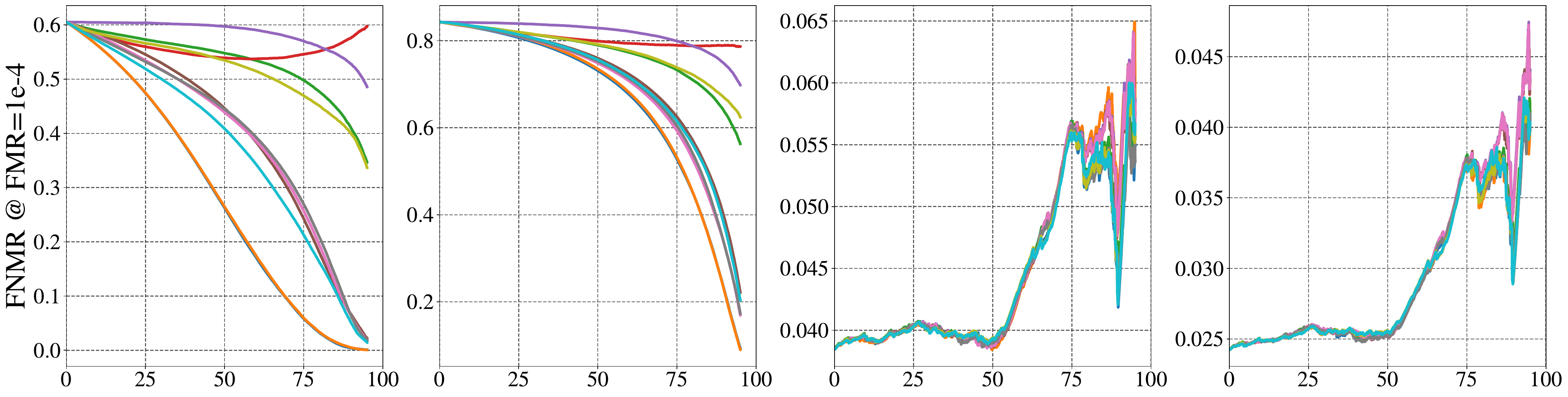}
  \includegraphics[width=\linewidth]{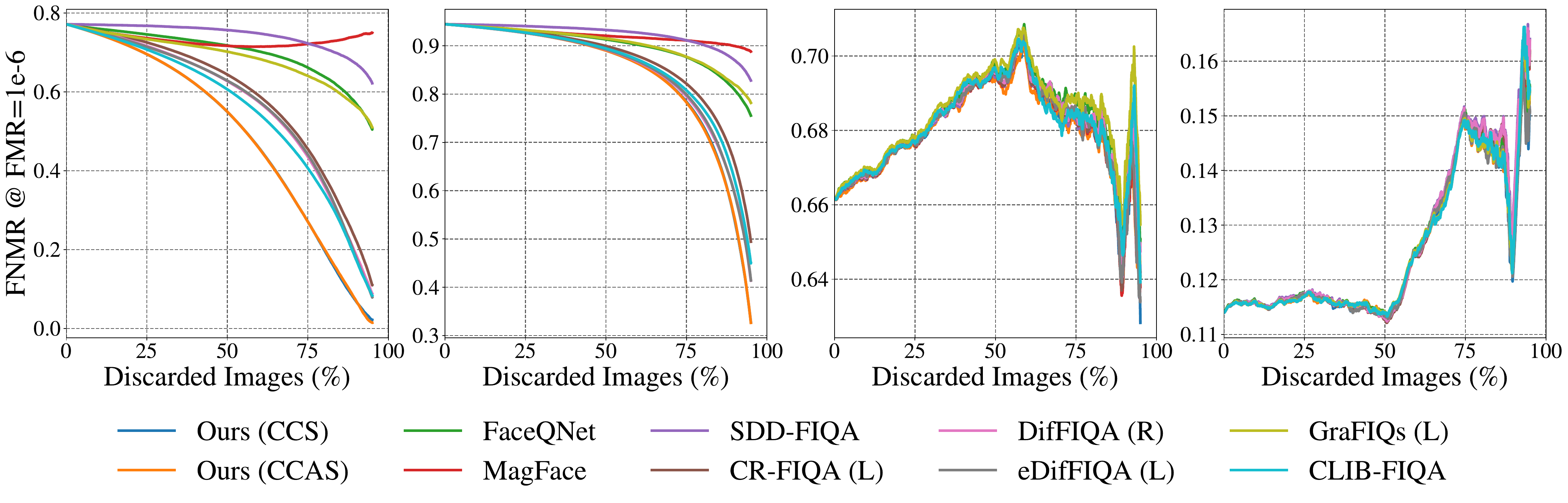}
  \vspace{-7mm}
  \caption{\textbf{ERC analysis at multiple operating points.} 
  Top: image-level ERC curves at a target FMR of $10^{-3}$. 
  Middle: curves at $10^{-4}$. 
  Bottom: curves at $10^{-6}$. 
  Curves closer to the bottom left indicate more effective trade-offs between discarding fraction and FNMR. 
  Corresponding AUCs and Spearman correlations are reported in Table~\ref{tab:erc_auc_results}. 
  For clarity, only the strongest variant of each method is shown.}
  \label{fig:erc_app}
\end{figure*}

Fig.~\ref{fig:erc_app} extends the main analysis in Section~\ref{sec:experiments:erc} by showing ERC curves at stricter target FMRs of $10^{-4}$ and $10^{-6}$. The trends remain consistent: CCS- and CCAS-based predictions achieve the highest Spearman correlations and lowest AUCs across all settings. The only exception is Swin at $10^{-4}$, where our method performs nearly identically to eDifFIQA (L). These results reinforce that recognizability-grounded supervision provides state-of-the-art alignment with recognition reliability, even under the strictest operating conditions.  

\end{document}